\documentclass[11pt,a4paper]{article}

\usepackage[utf8]{inputenc}
\usepackage{amsmath}
\usepackage{amssymb}
\usepackage{graphicx}
\usepackage[margin=1in]{geometry}
\usepackage{hyperref}
\usepackage{caption}
\usepackage{booktabs} 
\usepackage{authblk} 

\hypersetup{
    colorlinks=true,
    linkcolor=blue,
    filecolor=magenta,      
    urlcolor=cyan,
    pdftitle={Real-Time Beach Litter Detection and Counting: A Comparative Analysis of RT-DETR Model Variants},
    pdfpagemode=FullScreen,
}

\title{\textbf{Real-Time Beach Litter Detection and Counting: A Comparative Analysis of RT-DETR Model Variants}}

\author[1]{Miftahul Huda*}
\author[1]{Arsyiah Azahra}
\author[1]{Putri Maulida Chairani}
\author[1]{Dimas Rizky Ramadhani}
\author[1]{Nabila Azhari}
\author[1]{Ade Lailani}

\affil[1]{Data Science Study Program, Faculty of Science, Sumatera Institute of Technology}

\date{} 

\begin{document}

\maketitle

\begin{abstract}
\noindent Coastal pollution is a pressing global environmental issue, necessitating scalable and automated solutions for monitoring and management. This study investigates the efficacy of the Real-Time Detection Transformer (RT-DETR), a state-of-the-art, end-to-end object detection model, for the automated detection and counting of beach litter. A rigorous comparative analysis is conducted between two model variants, RT-DETR-Large (RT-DETR-L) and RT-DETR-Extra-Large (RT-DETR-X), trained on a publicly available dataset of coastal debris. The evaluation reveals that the RT-DETR-X model achieves marginally superior accuracy, with a mean Average Precision at 50\% IoU (mAP@50) of 0.816 and a mAP@50-95 of 0.612, compared to the RT-DETR-L model's 0.810 and 0.606, respectively. However, this minor performance gain is realized at a significant computational cost; the RT-DETR-L model demonstrates a substantially faster inference time of 20.1 ms versus 34.5 ms for the RT-DETR-X. The findings suggest that the RT-DETR-L model offers a more practical and efficient solution for real-time, in-field deployment due to its superior balance of processing speed and detection accuracy. This research provides valuable insights into the application of advanced Transformer-based detectors for environmental conservation, highlighting the critical trade-offs between model complexity and operational viability.

\vspace{1em}
\end{abstract}

\section{Introduction}
The accumulation of waste along coastlines has emerged as one of the most urgent environmental crises worldwide. This debris, particularly plastic, not only degrades the aesthetic quality of natural landscapes but also inflicts severe damage upon marine ecosystems. Plastic waste, which constitutes approximately 60\% of all marine debris, persists in the environment for hundreds of years and is often ingested by marine fauna, leading to detrimental health effects that cascade through the food web and can ultimately impact human health \cite{jambeck2015, worldbank2022}. The scale of this problem, with a significant portion of waste originating from coastal human activities, necessitates effective and scalable methods for monitoring and removal.

Traditional methods for managing beach litter rely on manual surveys and clean-up efforts, which are inherently labor-intensive, time-consuming, and costly \cite{liu2021, borthakur2019}. These limitations hinder the ability to conduct frequent, large-scale assessments required for effective environmental management. In response, the fields of computer vision and artificial intelligence (AI) have presented powerful alternatives for automating the surveillance and data collection process. Automated systems can analyze vast amounts of visual data from sources like drones or fixed cameras, enabling rapid, accurate, and consistent identification and quantification of litter, thereby overcoming the logistical challenges of manual approaches.

The technological backbone of these automated systems has evolved significantly. Early successes were driven by Convolutional Neural Network (CNN) based object detectors, such as the YOLO (You Only Look Once) and SSD (Single Shot MultiBox Detector) families, which prioritized high-speed processing. A paradigm shift occurred with the introduction of Transformer architectures to computer vision, culminating in models like the Detection Transformer (DETR) \cite{carion2020}. DETR introduced an elegant end-to-end approach that eliminated the need for many hand-designed components like non-maximum suppression. However, the high computational complexity of its Transformer encoder made the original DETR model ill-suited for real-time applications where low latency is critical \cite{he2021}.

This paper addresses the need for both high accuracy and real-time performance in environmental monitoring by investigating the Real-Time Detection Transformer (RT-DETR) \cite{zhao2023}. RT-DETR is a state-of-the-art architecture designed to overcome the computational bottlenecks of its predecessors while maintaining a robust, end-to-end detection pipeline. This study conducts a rigorous comparison of two RT-DETR model variants—Large (RT-DETR-L) and Extra-Large (RT-DETR-X) applied specifically to the domain of beach debris detection. By systematically evaluating the trade-offs between model complexity, detection accuracy, and inference speed, this study aims to identify the most viable architecture for practical, in-field deployment in environmental conservation efforts.

\section{Methodology}

\subsection{Dataset}
The dataset utilized in this study is the "Beach Waste Dataset" provided by Monash and publicly available on Roboflow under a Creative Commons Attribution 4.0 International (CC BY 4.0) license \cite{monash2024, creativecommons2024}. The dataset is composed of images depicting various forms of litter commonly found on coastlines. It is partitioned into a training set of 2,675 images and a validation set of 841 images. The annotations cover seven distinct object categories: "Bottle," "Clothes," "Metal," "Plastic," "Rope," "Styrofoam," and "Wood". Each object instance is annotated with a bounding box defined by its normalized center coordinates ($x, y$), width, and height, along with its corresponding class label.

A critical limitation of this dataset is that the validation set does not contain any instances of the "Wood" and "Clothes" categories. Consequently, while the models were trained on all seven classes, their performance on these two specific classes could not be quantitatively evaluated using the provided validation split. The reported performance metrics are therefore calculated based only on the five classes present in the validation data.

\subsection{RT-DETR Architecture}
\begin{figure}[h!]
    \centering
    \includegraphics[width=\textwidth]{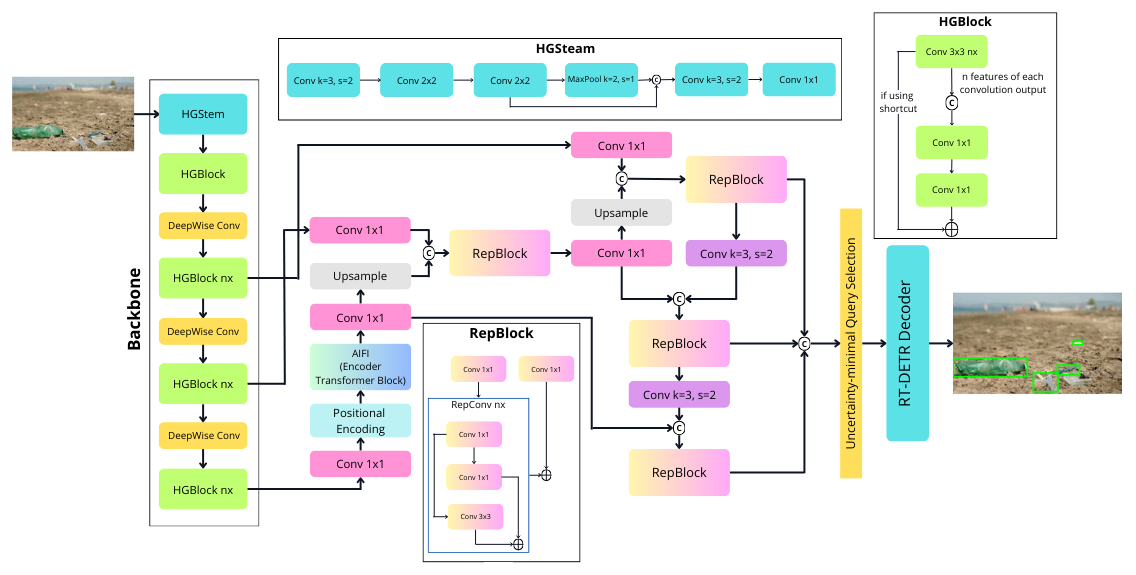}
    \caption{RT-DETR Architecture. Features with three different scales generated by the backbone are processed through the Efficient Hybrid Encoder. Features from the last layer pass through a Transformer layer, which applies an Attention-based Intra-scale Feature Interaction (AIFI) mechanism for channel-wise attention. The three features are combined, and then Uncertainty-minimal Query Selection is applied to choose features from the encoder as object queries for the decoder. Finally, the decoder with its heads optimizes the object queries to produce object categories and bounding box coordinates.}
    \label{fig:architecture}
\end{figure}

RT-DETR is an end-to-end object detector that features a hybrid design, integrating a CNN backbone for efficient multi-scale feature extraction with a Transformer-based encoder-decoder for object detection and classification \cite{zhao2023, ultralytics2024}. The detailed architecture is depicted in Figure \ref{fig:architecture}.

\begin{itemize}
    \item \textbf{Efficient Hybrid Encoder:} The core of RT-DETR's real-time capability lies in its Efficient Hybrid Encoder. This component is specifically designed to process the multi-scale feature maps generated by the CNN backbone in a computationally efficient manner. It employs mechanisms like Attention-based Intra-scale Feature Interaction (AIFI) to model relationships within features at the same scale, avoiding the high cost of global attention across all scales.
    \item \textbf{Uncertainty-minimal Query Selection:} To improve accuracy, RT-DETR incorporates an Uncertainty-minimal Query Selection mechanism. Instead of using fixed or learned object queries, this module intelligently selects the most informative features from the encoder's output to serve as initial object queries for the decoder. The selection is optimized during training by minimizing the uncertainty between localization and classification predictions, ensuring that the decoder focuses on high-quality, relevant proposals.
    \item \textbf{Backbone Variants:} This study compares two RT-DETR variants: Large (L) and Extra-Large (X). The primary architectural distinction between them lies in the depth and complexity of the CNN backbone, specifically in the number of HGBlocks used at different stages \cite{ultralytics2024, chollet2017}. The RT-DETR-L model employs a sequence of one, three, and one HGBlocks, whereas the RT-DETR-X model uses a deeper configuration of two, five, and two HGBlocks, respectively. This increased depth in the XLarge model results in a greater number of parameters and higher computational requirements (FLOPs).
\end{itemize}

\subsection{Loss Function Formulation}
The RT-DETR model outputs a fixed set of $N$ predictions, where $N$ is the number of object queries in the decoder. To handle the mismatch between the fixed number of predictions and the variable number of ground-truth objects in an image, a bipartite matching process is required before computing the final loss \cite{carion2020}.

\paragraph{Hungarian Matching} The optimal assignment between predicted and ground-truth bounding boxes is found using the Hungarian algorithm. This algorithm seeks to find a permutation of $N$ elements, $\hat{\sigma}$, that minimizes the total matching cost \cite{ultralytics2024, lin2018}:
\begin{equation}
\hat{\sigma} = \arg\min_{\sigma \in S_{N}} \sum_{i=1}^{N} \mathcal{L}_{\text{match}}(y_{i}, \hat{y}_{\sigma(i)})
\end{equation}
The matching cost, $\mathcal{L}_{\text{match}}$, is a linear combination of a classification loss ($\mathcal{L}_{cls}$), an L1 bounding box regression loss ($\mathcal{L}_{bbox}$), and a Generalized Intersection over Union loss ($\mathcal{L}_{giou}$). These components are defined as follows, transcribed exactly from the source material \cite{ultralytics2024, lin2018, reza2019}:
\begin{equation}
\mathcal{L}_{cls} = \alpha(1-\hat{p})^{\gamma}(-\log(\hat{p})) - (1-\alpha)\hat{p}^{\gamma}(-\log(1-\hat{p}))
\end{equation}
\begin{equation}
\mathcal{L}_{bbox} = \sum_{i=1}^{N} |\hat{b}_{i} - b_{i}|
\end{equation}
\begin{equation}
\mathcal{L}_{giou} = 1 - \text{GIoU}(b_{i}, \hat{b}_{i})
\end{equation}
\begin{equation}
\mathcal{L}_{\text{match}} = \lambda_{cls}L_{cls} + \lambda_{bbox}L_{bbox} + \lambda_{giou}L_{giou}
\end{equation}
The hyperparameters for the matching cost are set to $\alpha=0.25$, $\gamma=2$, $\lambda_{cls}=1$, $\lambda_{bbox}=5$, and $\lambda_{giou}=2$.

\paragraph{Final Training Loss} Once the optimal matching is established, the total training loss $\mathcal{L}$ is computed as a weighted sum of the individual loss components over the matched pairs. The total loss is defined as follows, transcribed exactly from the source material \cite{zhao2023, ultralytics2024, lin2018, zhang2021, reza2019}:
\begin{equation}
\begin{split}
\mathcal{L} = \ &\frac{\lambda_{giou}}{N} \sum_{i=1}^{N} \text{GIoU}(\hat{b}_{i}, b_{i}) + \frac{\lambda_{l1}}{N} \sum_{i=1}^{N} |\hat{b}_{i} - b_{i}| \\
&+ \frac{\lambda_{cls}}{N} \sum_{i=1}^{N} \sum_{c=1}^{C} \left[ (q_{i,c} \log(\hat{p}_{i,c}) + (1-q_{i,c}) \log(1-\hat{p}_{i,c})) (\alpha \hat{p}_{i,c}^{\gamma}(1-p_{i,c}) + q_{i,c} p_{i,c}) \right]
\end{split}
\end{equation}
Here, $q$ represents the IoU between a predicted and ground-truth box. The loss weighting hyperparameters are maintained from the matching cost: $\lambda_{giou}=2$, $\lambda_{l1}=5$ (representing L1 loss), and $\lambda_{cls}=1$ \cite{ultralytics2024}.

\subsection{Training and Evaluation Protocol}
Both the RT-DETR-L and RT-DETR-X models were trained using pre-trained weights and the source code from the Ultralytics repository \cite{ultralytics2024}. The training was conducted on a single NVIDIA P100 GPU with a batch size of 12. The AdamW optimizer was used for model training \cite{loshchilov2019}.

Model performance was evaluated using standard object detection metrics: Precision, Recall, mean Average Precision at an IoU threshold of 0.5 (mAP@50), and mean Average Precision averaged over IoU thresholds from 0.5 to 0.95 in steps of 0.05 (mAP@50-95), which aligns with the COCO evaluation protocol.

\section{Experimental Results and Analysis}

\subsection{Computational Performance and Efficiency Trade-offs}
A primary objective of this study was to quantify the trade-off between model complexity and computational efficiency. Table \ref{tab:complexity} presents a comparison of the parameter count, Giga Floating Point Operations (GFLOPs), and inference time for the two model variants.

\begin{table}[h!]
\centering
\caption{Model Complexity and Inference Speed Comparison.}
\label{tab:complexity}
\begin{tabular}{@{}lccc@{}}
\toprule
\textbf{Model} & \textbf{Parameters (M)} & \textbf{GFLOPs} & \textbf{Inference Time (ms)} \\ \midrule
RT-DETR-Large & 32.9 & 108.0 & 20.1 \\
RT-DETR-XLarge & 67.3 & 234.4 & 34.5 \\ \bottomrule
\end{tabular}
\end{table}

The data clearly illustrates the significant increase in computational demand associated with the RT-DETR-X model. It contains more than double the number of parameters (2.05x) and requires over twice the computational operations (2.17x GFLOPs) compared to the RT-DETR-L model \cite{desislavov2023}. This increased complexity directly translates to a 71.6\% longer inference time. This difference is critical for real-world deployment; an inference time of 20.1 ms allows the RT-DETR-L model to process video at approximately 50 frames per second (FPS), well within the requirements for most real-time applications. In contrast, the 34.5 ms latency of the RT-DETR-X model corresponds to a much lower throughput of around 29 FPS.

\subsection{Quantitative Detection Performance}
The detection accuracy of both models was evaluated on the validation set. Table \ref{tab:overall_perf} summarizes the overall performance, while Table \ref{tab:per_class_perf} provides a detailed per-class breakdown.

\begin{table}[h!]
\centering
\caption{Overall Detection Performance Comparison.}
\label{tab:overall_perf}
\begin{tabular}{@{}lcccc@{}}
\toprule
\textbf{Model} & \textbf{Precision} & \textbf{Recall} & \textbf{mAP@50} & \textbf{mAP@50-95} \\ \midrule
RT-DETR-Large & 0.848 & 0.744 & 0.810 & 0.606 \\
RT-DETR-XLarge & 0.868 & 0.743 & 0.816 & 0.612 \\ \bottomrule
\end{tabular}
\end{table}

\begin{table}[h!]
\centering
\caption{Per-Class Performance Breakdown.}
\label{tab:per_class_perf}
\resizebox{\textwidth}{!}{%
\begin{tabular}{@{}llccccc@{}}
\toprule
\textbf{Class} & \textbf{Model} & \textbf{Instances} & \textbf{Precision} & \textbf{Recall} & \textbf{mAP@50} & \textbf{mAP@50-95} \\ \midrule
\textbf{Bottle} & Large & 874 & 0.906 & 0.814 & 0.872 & 0.638 \\
 & XLarge & & 0.908 & 0.827 & 0.871 & 0.645 \\ \addlinespace
\textbf{Metal} & Large & 458 & 0.882 & 0.852 & 0.878 & 0.648 \\
 & XLarge & & 0.925 & 0.832 & 0.882 & 0.658 \\ \addlinespace
\textbf{Plastic} & Large & 603 & 0.838 & 0.776 & 0.819 & 0.636 \\
 & XLarge & & 0.847 & 0.774 & 0.813 & 0.637 \\ \addlinespace
\textbf{Rope} & Large & 205 & 0.709 & 0.512 & 0.619 & 0.409 \\
 & XLarge & & 0.796 & 0.478 & 0.640 & 0.413 \\ \addlinespace
\textbf{Styrofoam} & Large & 418 & 0.904 & 0.765 & 0.861 & 0.697 \\
 & XLarge & & 0.863 & 0.801 & 0.871 & 0.705 \\ \bottomrule
\end{tabular}%
}
\end{table}

The results in Table \ref{tab:overall_perf} show that the RT-DETR-X model provides only a marginal improvement in overall detection accuracy. The mAP@50 increases by just 0.6 percentage points (from 0.810 to 0.816), and the more stringent mAP@50-95 metric shows an identical 0.6 percentage point gain. This negligible increase in accuracy appears highly disproportionate to the over 70\% increase in inference latency and doubling of computational cost.

\subsection{Qualitative Assessment and Error Analysis}
\begin{figure}[h!]
    \centering
    \includegraphics[width=0.6\textwidth]{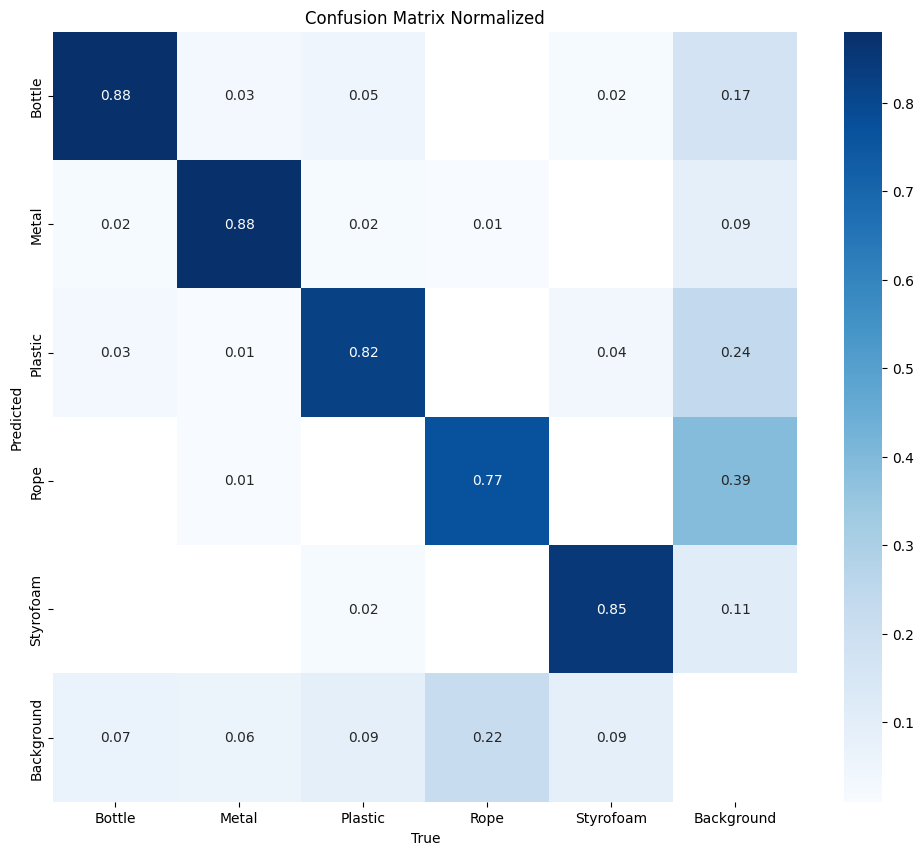}
    \caption{Normalized confusion matrix for the RT-DETR-Large model, showing significant confusion between the "Rope" and "Background" classes.}
    \label{fig:cm_large}
\end{figure}

\begin{figure}[h!]
    \centering
    \includegraphics[width=0.6\textwidth]{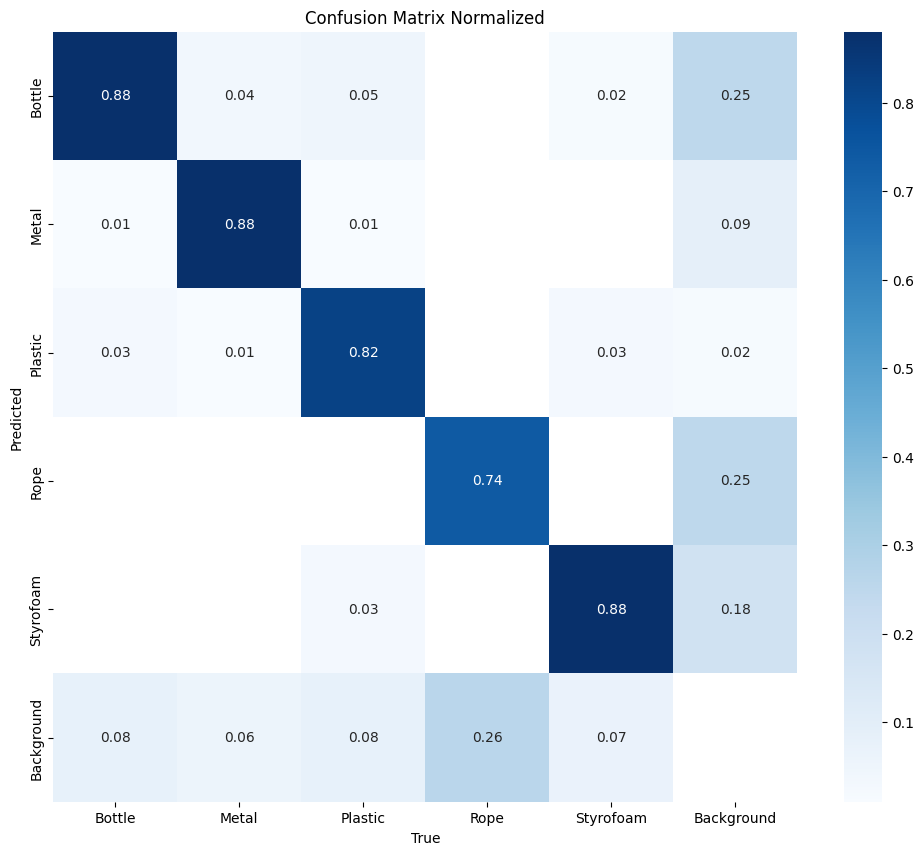}
    \caption{Normalized confusion matrix for the RT-DETR-XLarge model. The "Rope" vs. "Background" confusion persists.}
    \label{fig:cm_xlarge}
\end{figure}

To further understand model behavior, a qualitative analysis was performed using normalized confusion matrices and sample inference images. The confusion matrices for both the Large (Figure \ref{fig:cm_large}) and XLarge (Figure \ref{fig:cm_xlarge}) models visually confirm the quantitative findings. A prominent off-diagonal value in both matrices indicates significant confusion between the "Rope" class and the "Background" class.

\begin{figure}[h!]
    \centering
    \includegraphics[width=\textwidth]{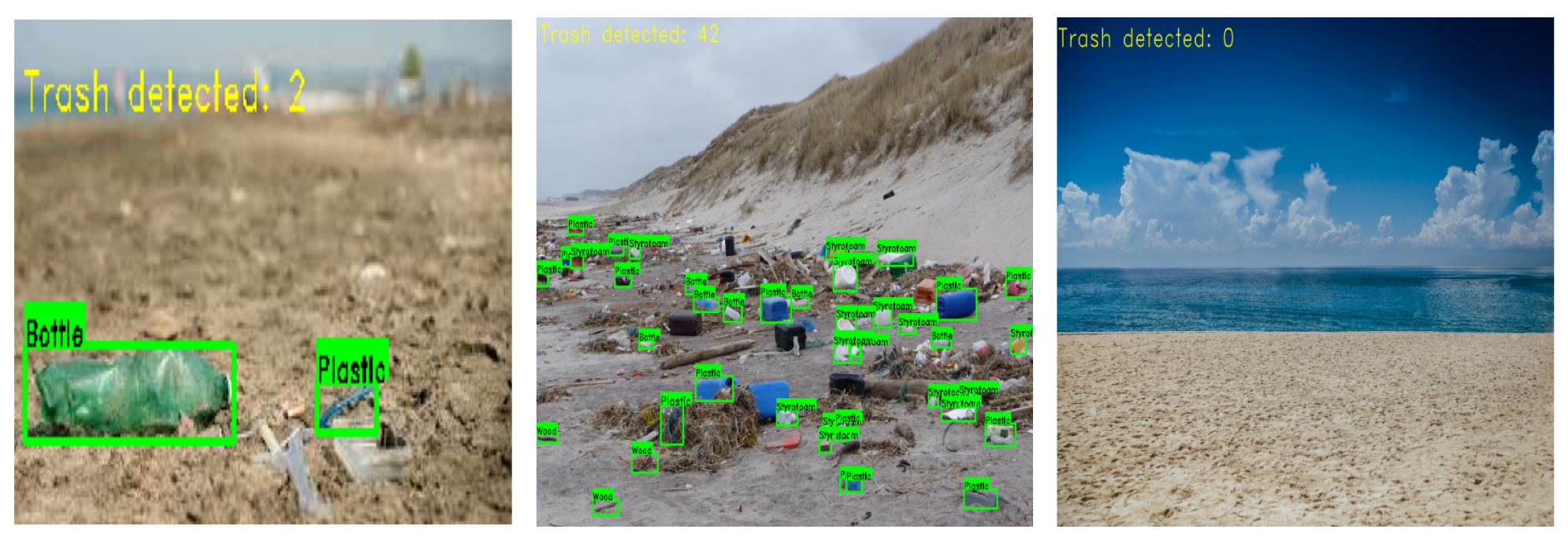}
    \caption{Sample predictions from the RT-DETR-Large model across different litter densities.}
    \label{fig:pred_large}
\end{figure}

\begin{figure}[h!]
    \centering
    \includegraphics[width=\textwidth]{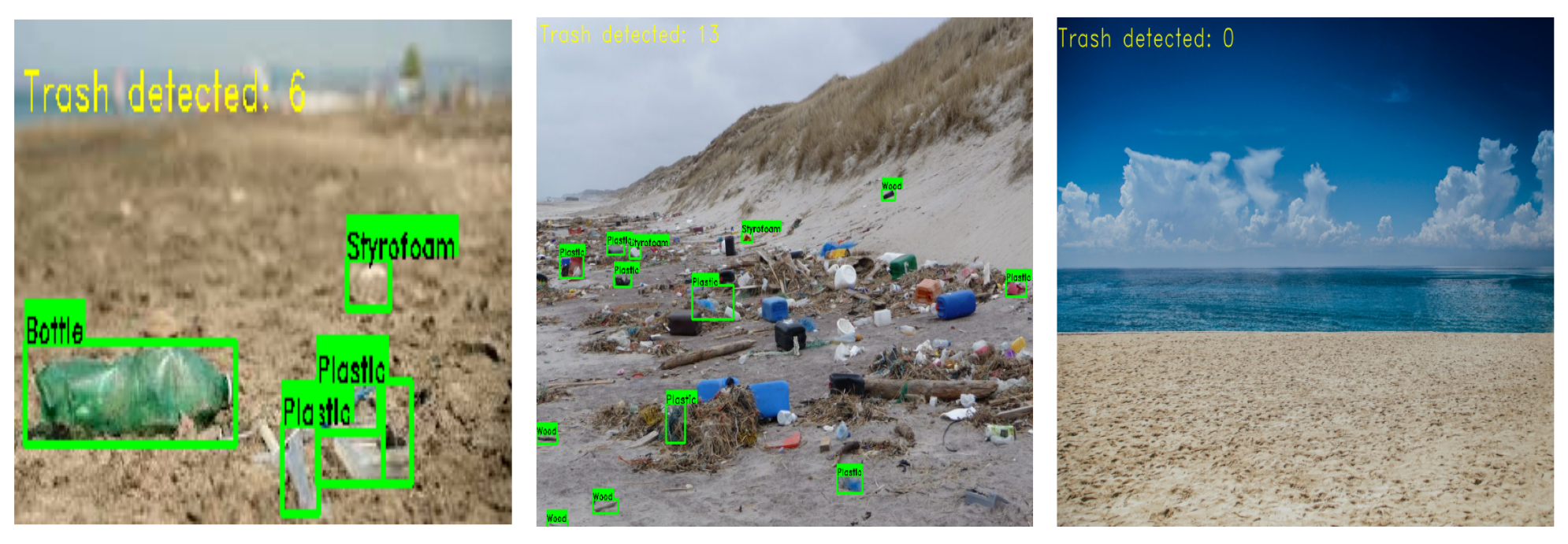}
    \caption{Sample predictions from the RT-DETR-XLarge model across different litter densities.}
    \label{fig:pred_xlarge}
\end{figure}

Generalization tests were conducted on sample images representing three common scenarios: low-density litter, high-density litter, and a clean beach (Figures \ref{fig:pred_large} and \ref{fig:pred_xlarge}). In a scene with few, distinct objects, the XLarge model demonstrated higher sensitivity. However, in a highly cluttered scene, the Large model detected 42 objects, while the XLarge model detected only 6, suggesting the larger model may have learned a tighter decision boundary, leading to higher precision but lower recall in complex scenes. On a clean beach, both models correctly predicted zero objects, demonstrating strong performance in avoiding false positives.

\section{Conclusion and Future Work}
This research demonstrates that the RT-DETR framework is a highly effective tool for the automated detection of beach litter. Through a comparative analysis, the RT-DETR-Large model was identified as the optimal architecture for practical deployment, offering a compelling balance between high detection accuracy and real-time processing speed. The RT-DETR-XLarge model, while marginally more accurate, is hindered by computational demands that make it less suitable for in-field applications.

The consistent underperformance on the "Rope" class underscores a fundamental challenge for standard bounding-box-based detectors with amorphous objects. Future work should proceed along several key directions. First, model optimization techniques such as quantization and pruning could be applied to the RT-DETR-L model to enhance suitability for deployment on resource-constrained edge devices. Second, the challenge of detecting amorphous objects could be addressed by exploring instance segmentation models. Finally, the development of a more comprehensive and geographically diverse dataset is crucial for building more robust and generalizable models for integration into a fully autonomous environmental management system.

\section*{Acknowledgments}
This work was supported in part by the Data Science Study Program at the Sumatera Institute of Technology. The authors extend their gratitude to Ade Lailani and Christyan Tamaro Nadeak for their valuable suggestions on the writing, presentation, and discussion of this research.


\end{document}